# Analytical Verification of Deep Neural Network Performance for Time-Synchronized Distribution System State Estimation

Behrouz Azimian, *Student Member*, *IEEE,* Shiva Moshtagh, *Student Member*, *IEEE*, Anamitra Pal, *Senior Member, IEEE*, and Shanshan Ma, *Member*, *IEEE*

*Abstract*— **Recently, we demonstrated success of a time-synchronized state estimator using deep neural networks (DNNs) for real-time unobservable distribution systems [1]. In this letter, we provide analytical bounds on the performance of that state estimator as a function of perturbations in the input measurements. It has already been shown that evaluating performance based on only the test dataset might not effectively indicate a trained DNN's ability to handle input perturbations. As such, we analytically verify robustness and trustworthiness of DNNs to input perturbations by treating them as mixed-integer linear programming (MILP) problems. The ability of batch normalization in addressing the scalability limitations of the MILP formulation is also highlighted. The framework is validated by performing time-synchronized distribution system state estimation for a modified IEEE 34-node system and a real-world large distribution system, both of which are incompletely observed by micro-phasor measurement units.**

## I. Introduction

DISTRIBUTION system state estimation (DSSE) utilizing micro-phasor measurement units (μPMUs) and deep neural networks (DNNs) is currently a topic of active research interest in the power systems community. This is because their combination can provide high-speed situational awareness in *real-time unobservable* distribution systems as already demonstrated in [1]-[2]. However, DNNs are vulnerable to perturbations in their inputs that lead to errors in their outputs, many of which are not captured during DNN training/validation [3]-[4]. This is because DNN hyper-parameters are chosen to minimize validation loss and not to be robust against input perturbations. In this letter, we establish formal guarantees of DNN performance to theoretically prove that for a bounded perturbation in the inputs, the errors in the outputs of a trained DNN are also bounded.

Providing performance guarantees to machine learning (ML) algorithms (DNN being a type of ML algorithm) is particularly important for power system problems as the electric power grid is a mission-critical system. In line with this realization, [5] and [6] have provided performance guarantees to ML algorithms for power system classification problems, while [7] focused on a power system regression problem. However, the ML algorithms investigated in [5]-[7] were shallow models. For example, [5] and [6] used support vector classification and decision trees, respectively, while [7] used linear regression and support vector regression. These shallow ML algorithms are not suitable for performing time-synchronized DSSE in modern distribution systems due to the severity of real-time unobservability and the increased uncertainty caused by behind-the-meter generation. Recently, a nontrivial certified lower bound of minimum adversarial distortion has been derived for a general class of ML problems involving DNNs in [8]-[11], and for a DNN-based *classification* problem of power systems in [12]. The methodology developed in [12] was extended in [13] to provide worst-case performance guarantees by integrating physics-based constraints directly into the trained DNN. However, the scarcity of μPMUs in the distribution system makes it impossible to analytically relate their measurements with every state. In summary, to the best of our knowledge, no prior work has investigated the verification of DNNs for power system regression problems in which the underlying analytical relation between inputs and outputs is unavailable.

In this letter, we exploit the piecewise linear nature of the rectified linear unit (ReLU) activation function, which is one of the most commonly used activation functions, to analytically examine robustness and trustworthiness of a trained DNN for performing time-synchronized DSSE in μPMU-unobservable distribution systems. We first express the ReLU operation *integrated with batch normalization* (BN) as a mixed-integer linear programming (MILP) problem. We then introduce two sets of verification formulations. For robustness verification, we show that *for a prespecified range of perturbation in the input, the deviation in the output from its reference value is guaranteed to lie within a bounded region*. For trustworthiness verification, we find *the minimum perturbation required in the input to generate a given error in the output*. Extensive simulations carried out using a modified IEEE 34-node distribution system demonstrate that the proposed formulations ensure robustness and trustworthiness of DNNs for time-synchronized DSSE. We have also tested our verification formulations on a real-world large distribution system to demonstrate their scalability and widespread applicability.

The salient contributions of this letter are as follows:

1. Providing bounds on the estimation error of a DNN- based time-synchronized distribution system state estimator given a bounded perturbation in the input measurements (robustness verification).

Manuscript received: June 21, 2023; accepted: October 15, 2023. Date of CrossCheck: XXX XX XXXX. Date of online publication: XXX XX, XXXX.

This work was supported in part by the Department of Energy under grants DE-AR-0001001 and DE-EE0009355, and the National Science Foundation (NSF) under the grant ECCS-2145063.

B. Azimian, S. Moshtagh, and A. Pal (corresponding author) are with the School of Electrical, Computer and Energy Engineering, Arizona State University, Tempe, AZ 85281 USA, (e-mail: bazimian@asu.edu; smoshta1@asu.edu; Anamitra.Pal@asu.edu).

S. Ma was with the School of Electrical, Computer and Energy Engineering, Arizona State University, Tempe, AZ 85281 USA. She is now with Quanta Technology, Raleigh, NC 27607, USA, (SMa@quanta-technology.com).

2. Quantifying the minimum perturbation in the input measurements required to create a given amount of error in the state estimates (trustworthiness verification).

3. Integrating BN with the verification formulations to improve scalability.

The rest of the letter is structured as follows. Section II explains the role of DNNs in time-synchronized DSSE. The robustness and trustworthiness formulations are developed in Section III. The results obtained are summarized and discussed in Section IV, while the conclusion is presented in Section V.

## II. Time-Synchronized DSSE using DNNs

In [1], we formulated a Bayesian approach to perform time-synchronized DSSE for systems that are incompletely observed by μPMUs. The resulting minimum mean squared error (MMSE) estimator minimized the estimation error of each state, $x_i$, for a given μPMU measurement vector, $\boldsymbol{z}$, as shown below,

$$\min_{\hat{x}_i(\cdot)} \mathbb{E}(\| x_i - \hat{x}_i(\boldsymbol{z}) \|^2) \Longrightarrow \hat{x}_i^*(\boldsymbol{z}) = \mathbb{E}(x_i|\boldsymbol{z}) \ \forall i \in [1, M] \quad (1)$$

where, $M$ denotes the total number of states to be estimated. The conditional expectation of (1) can be expressed in terms of the joint probability density of $x_i$ and $\boldsymbol{z}$, as shown in (2).

$$\mathbb{E}(x_i|\boldsymbol{z}) = \int_{-\infty}^{+\infty} x_i p(x_i|\boldsymbol{z}) dx_i = \int_{-\infty}^{+\infty} x_i \frac{p(x_i, \boldsymbol{z})}{p(\boldsymbol{z})} dx_i \quad (2)$$

For μPMU-unobservable systems, the probability density function (PDF) between μPMU data and all the voltage phasors (states) is unknown or impossible to specify, making direct computation of $\hat{x}_i^*(z)$ intractable. Even if the underlying joint PDF is known, finding a closed-form solution to (2) can be hard. The DNN's role is to approximate the MMSE estimator as it has excellent approximation capabilities [14]; i.e., the DNN for DSSE finds a mapping function that relates $x_i$ and $\boldsymbol{z}$.

## III. Proposed Formulations

A well-trained DNN that gives satisfactory response for unseen test data cannot necessarily ensure similar performance for all possible combinations of its inputs. This brings into question the rationale of using DNNs for decision-making in mission-critical systems. The goal of this letter is to address this concern through verification-based methodologies and build credibility of DNNs for time-synchronized DSSE.

### A. Reformulating ReLU function with BN based on MILP

The DNN used in this analysis is a fully-connected feed-forward neural network with $K$ hidden layers integrated with BN, each having $N$ neurons, as shown in Fig. 1. The inputs and outputs are denoted by $\boldsymbol{z} \in \mathbb{R}^{N_0}$ and $\boldsymbol{x} \in \mathbb{R}^M$, respectively, where, $N_0 \ll M$ for unobservable systems. The hidden layers, denoted by $\boldsymbol{h}_k \in \mathbb{R}^N$, are equipped with ReLU activation function. The input to the ReLU function is a linear transformation of the output of the previous layer denoted by $\hat{\boldsymbol{h}}_k \in \mathbb{R}^N$. The output of each neuron in every hidden layer is sent to a BN operator. Hence, for each layer, we have:

$$\hat{\boldsymbol{h}}_k = \boldsymbol{W}_k \boldsymbol{BN}_{k-1} + \boldsymbol{b}_k \qquad \forall k \in [1, K] \quad (3)$$

$$h_k^n = \max\left(\hat{h}_k^n, 0\right) \qquad \forall k \in [1, K], \forall n \in [1, N] \quad (4)$$

$$BN_k^n = \frac{\gamma_k^n (h_k^n - \mu_k^n)}{\sqrt{var_k^n + \varepsilon}} + \beta_k^n \qquad \forall k \in [1, K], \forall n \in [1, N] \quad (5)$$

where, $\boldsymbol{W}$ is the weight matrix, $\boldsymbol{b}$ is the bias vector, $\varepsilon$ is a small configurable constant, and $\gamma$ and $\beta$ are scaling and offset factors, respectively. The values of these hyperparameters are obtained during the training process. Note that in (3), $\boldsymbol{BN}_0 = \boldsymbol{z}$, while in (5), $\mu$ and $var$ denote moving average and variance of the batches seen during the training process. These are non-trainable variables that are updated each time the layer is called during the training process based on the given batch. As such,

$$\mu = \eta\mu + (1 - \eta)\mathbb{E}(\text{batch}) \quad (6)$$

$$var = \eta var + (1 - \eta)(\mathbb{E}(\text{batch}^2) - \mathbb{E}^2(\text{batch})) \quad (7)$$

where, $\eta$ is a configurable constant called momentum. In accordance with [8]-[9], the ReLU function defined in (4) is reformulated as a MILP problem. Defining the binary variable, $r_k^n \in \{0, 1\}$, for all hidden layers $k \in [1, K]$, and each neuron $n \in [1, N]$, we can rewrite the ReLU function as:

$$\begin{matrix} h_k^n = \max\left(\hat{h}_k^n, 0\right) \\ r_k^n \in \{0, 1\} \end{matrix} \Longrightarrow \begin{cases} h_k^n \le \hat{h}_k^n - \hat{h}_{k_{\min}}^n (1 - r_k^n) \\ h_k^n \ge \hat{h}_k^n \\ h_k^n \le \hat{h}_{k_{\max}}^n r_k^n \\ h_k^n \ge 0 \end{cases} \quad (8)$$

where, $r_k^n$ indicates whether the corresponding ReLU neuron is active (= 1) or inactive (= 0), and $\hat{h}_{k_{\min}}^n$ and $\hat{h}_{k_{\max}}^n$ are the upper and lower bounds of the ReLU function, respectively. The two bounds are calculated using the following equations:

$$\hat{\boldsymbol{h}}_{k_{\max}} = \max(\boldsymbol{W}_k, 0) \max\left(\boldsymbol{h}_{k_{\max}}, 0\right) + \min(\boldsymbol{W}_k, 0) \max\left(\boldsymbol{h}_{k_{\min}}, 0\right) + \boldsymbol{b}_k \quad (9a)$$

$$\hat{\boldsymbol{h}}_{k_{\min}} = \max(\boldsymbol{W}_k, 0) \max\left(\boldsymbol{h}_{k_{\min}}, 0\right) + \min(\boldsymbol{W}_k, 0) \max\left(\boldsymbol{h}_{k_{\max}}, 0\right) + \boldsymbol{b}_k \quad (9b)$$

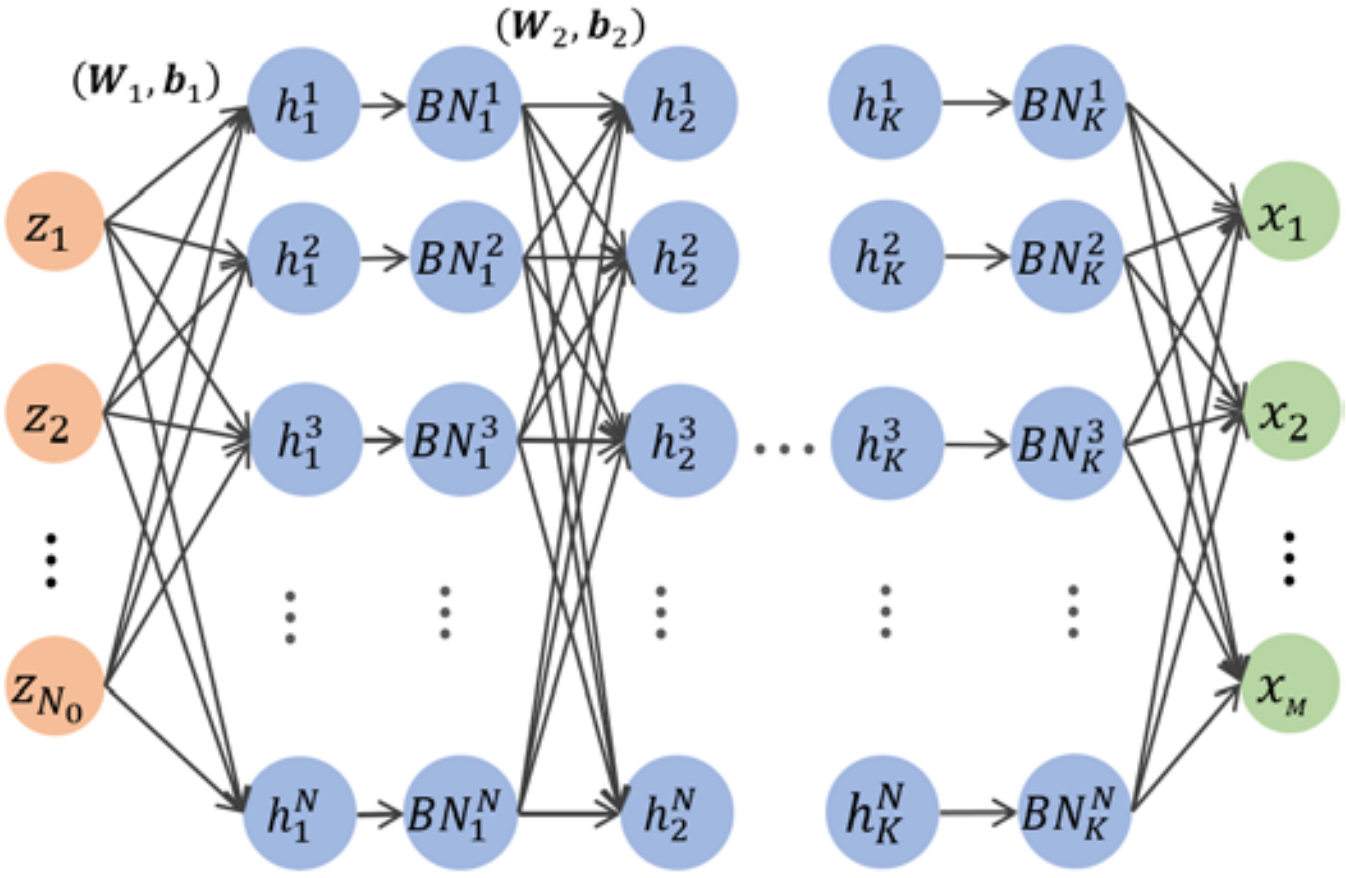

Fig. 1. DNN architecture with $K$ hidden layers, each having $N$ neurons; $x$ denotes the outputs obtained from the inputs, $z$.

For example, for the first hidden layer, $\boldsymbol{h}_{k_{\max}}$ and $\boldsymbol{h}_{k_{\min}}$ will correspond to the normalized inputs, implying that $\boldsymbol{h}_{1_{\max}} = 1$ and $\boldsymbol{h}_{1_{\min}} = 0$. This means that the bounds on the first hidden layer will be, $\hat{\boldsymbol{h}}_{1_{\max}} = \max(\boldsymbol{W}_1, 0) + \boldsymbol{b}_1$ and $\hat{\boldsymbol{h}}_{1_{\min}} = \min(\boldsymbol{W}_1, 0) + \boldsymbol{b}_1$, respectively. The bounds for the remaining layers can be obtained by applying (9a) and (9b) sequentially.

Lastly, $\boldsymbol{x}$ is calculated based on the linear transformation of the output of the BN operator in the last hidden layer, $\boldsymbol{BN}_K$.

In the proposed framework, one integer variable is assigned to each neuron for the linearization of the ReLU function. Hence, the number of integer variables quickly increases as deeper and wider DNNs are used. BN enables us to significantly reduce the number of integer variables while maintaining the high accuracy of the DNN. This is due to three main reasons [15]-[16]. Firstly, the BN layer combats the vanishing gradient problem by normalizing activations, which prevents creation of very small gradients during training. Secondly, it counters internal covariate shift by maintaining a consistent input distribution in each layer, which promotes stable and efficient training. Lastly, it acts as a form of regularization by introducing noise, which helps prevent overfitting and enhancing the DNN's ability to generalize to new data. These abilities of the BN layer help reduce the size of the DNN without compromising its accuracy, resulting in a direct improvement in the efficacy of the proposed formulations (see Section IV.C.1 for additional explanation on scalability).

### B. *Formulating Robustness for Regression Problems*

In this sub-section, we examine the robustness of DNNs for time-synchronized DSSE. Given a prespecified bounded perturbation in the input that deviates it from the actual value (reference), a DNN will be deemed robust if the output deviation is guaranteed to be within an acceptable threshold. This is pictorially depicted in Fig. 2 for a two-input, one-output DNN. The distortion-free input measurement, $\boldsymbol{z}_{ref}$, can be perturbed in either or both dimensions, with the maximum perturbation limit indicated by the black rectangle. For the time-synchronized DSSE application, the limit is specified in terms of permissible error in μPMU measurements (denoted by $\boldsymbol{\alpha}$). Now, for any randomly selected perturbed sample obtained during training/validation (purple dot), there can be a perturbed *adversarial* sample (black dot) encountered during testing that causes maximum error in the output (red oval). The goal of the robustness analysis is to quantify this maximum output error given the prespecified input perturbation limit, $\boldsymbol{\alpha}$. The following verification formulations are proposed to this end:

$$\max_{x} \left\| \boldsymbol{x} - \boldsymbol{x}_{ref} \right\|_p \tag{10a}$$

$$\text{s.t.} \quad \left\| \boldsymbol{z} - \boldsymbol{z}_{ref} \right\|_p \leq \boldsymbol{\alpha} \tag{10b}$$

$$(3), (5), (8) \tag{10c}$$

In (10), $\boldsymbol{x}_{ref}$ is a known output (e.g., true value of the state), $p \geq 1$ is an appropriate norm, (3) and (5) are DNN and BN constraints, and (8) denotes the linearized constraints of the ReLU function. By its very definition, (10) finds the maximum perturbation in the outputs corresponding to an input perturbation that is bounded by $\boldsymbol{\alpha}$. Consequently, it provides formal guarantees of robustness of a DNN for any regression problem involving ReLU activation function.

To prove the robustness of the trained DNN for all possible input combinations, we choose $p$ to be the infinity norm as it ensures that the DNN error is bounded throughout. For infinity norm maximization, we rewrite the objective function of (10a) as: $\max_{x} \max \left\{ \left| x_1 - x_{1_{ref}} \right|, \ldots, \left| x_M - x_{M_{ref}} \right| \right\}$. Next, we convert the overall maximization problem to one maximization and one minimization problem for each state. Finally, we pick the maximum absolute value between the two as shown below:

$$\max \left\{ \left| \max_{x_i} (x_i - x_{i_{ref}}) \right|, \left| \min_{x_i} (x_i - x_{i_{ref}}) \right| \right\} \; \forall i \in [1, M] \tag{11a}$$

$$\text{s.t.} \quad -\alpha_j \leq z_j - z_{j_{ref}} \leq \alpha_j \quad \forall j \in [1, N_0] \tag{11b}$$

$$(3), (5), (8) \tag{11c}$$

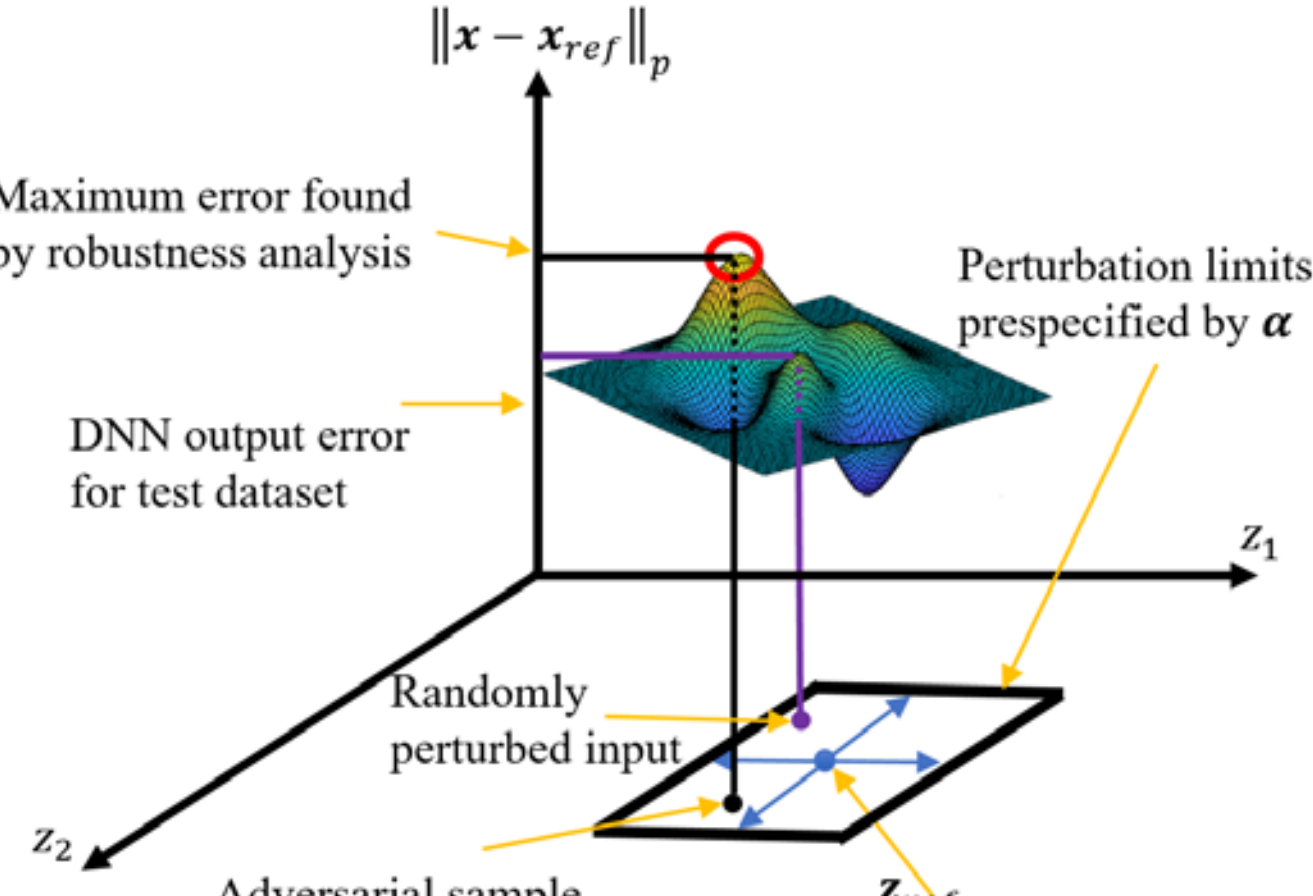

Fig. 2. Robustness analysis finds the adversarial sample for an operating condition described by $\boldsymbol{z}_{ref}$.

### C. *Formulating Trustworthiness for Regression Problems*

In this sub-section, we present a formulation for analyzing trustworthiness of a DNN trained for regression problems. If a perturbation in the input vector is denoted by $\boldsymbol{\delta}$, then the objective of trustworthiness analysis is to determine the smallest input perturbation (i.e., $\min(\boldsymbol{\delta}) = \boldsymbol{\delta}_{\min}$) that can create erroneous results exceeding a threshold, $\beta$, in the output. Afterwards, we compare the resulting $\boldsymbol{\delta}_{\min}$ with the actual level of perturbation allowed in the given application. For the time-synchronized DSSE problem, this will be the permissible error in μPMU measurements, specified by $\boldsymbol{\alpha}$. If $\boldsymbol{\delta}_{\min}$ consistently surpasses $\boldsymbol{\alpha}$, we can have trust in our trained DNN's ability to provide erroneous estimates that are always within $\beta$.

The above-mentioned logic is pictorially depicted in Fig. 3 for a two-input, one-output DNN. In the figure, the blue arrow represents the smallest input perturbation, $\boldsymbol{\delta}_{\min}$, which yields an error of $\beta$ in the output, while the green arrow represents $\boldsymbol{\alpha}$. As long as the blue arrow is longer than the green arrow for all scenarios, we can say with certainty that the estimation error will never surpass $\beta$. To find $\boldsymbol{\delta}_{\min}$, the following verification formulations are proposed.

$$\min \boldsymbol{\delta} \tag{12a}$$

$$\left\| \boldsymbol{z} - \boldsymbol{z}_{ref} \right\|_p \leq \boldsymbol{\delta} \tag{12b}$$

$$\text{s.t.} \quad \left\| \boldsymbol{x} - \boldsymbol{x}_{ref} \right\|_p \geq \beta \tag{12c}$$

$$(3), (5), (8) \tag{12d}$$

Eq. (12) is solved in a manner similar to how (11) was solved in the previous sub-section.

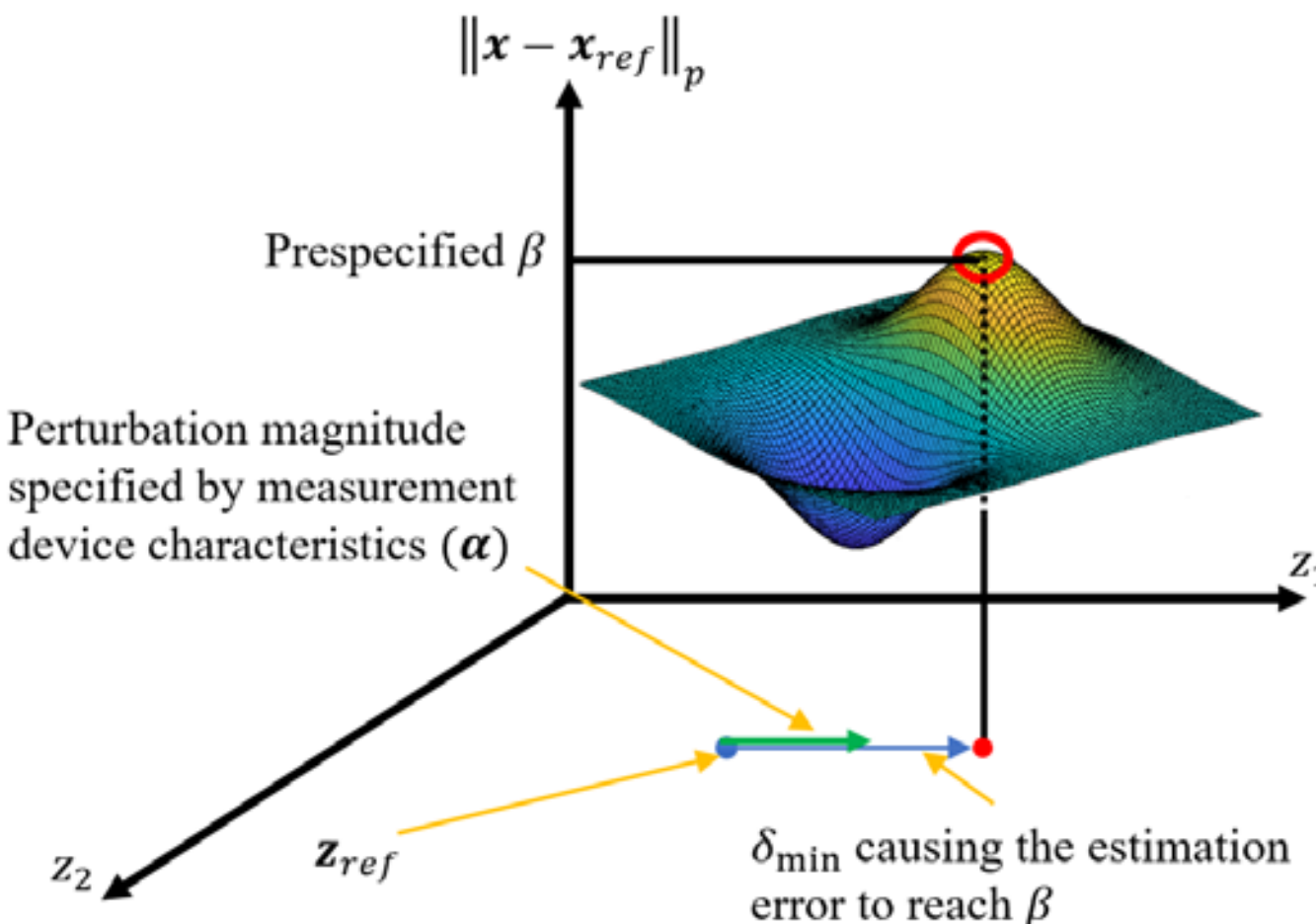

Fig. 3. Trustworthiness analysis finds a $\delta_{min}$ that yields a given error of $\beta$ in the output for each operating condition described by $\boldsymbol{z}_{ref}$.

### D. Data Preparation and Implementation of the Proposed Formulations

The DNN described in Section II is trained on historical smart meter data in the offline learning stage and tested using µPMU measurements in the online execution stage [1]. As the proposed verification formulations provide guarantees to the performance of this trained DNN, it is important to ensure that the learning and execution are done properly. For example, bad or missing data present in the smart meter measurements must be corrected; this was done by employing the data cleaning procedures described in [17]. Similarly, a Wald test-based bad data detection and correction procedure [18] was used to identify bad/missing µPMU measurements in real-time. These two procedures ensured that inaccurate or incomplete data do not limit the performance of the proposed approach.

To implement the proposed verification formulations, the following steps were performed:

Step 1: Cleaned smart meter data was used to create a big dataset by solving many power flows. The voltage phasors obtained from the power flow solution were saved as $\boldsymbol{x}_{ref}$. The voltage and current phasors corresponding to µPMU locations were saved as $\boldsymbol{z}_{ref}$. $\boldsymbol{z}$ was obtained from $\boldsymbol{z}_{ref}$ by using an appropriate perturbation limit, $\boldsymbol{\alpha}$.

Step 2: The big dataset was split into training and testing, and the former was used to train a DNN that finds a mapping function that relates $\boldsymbol{x}_{ref}$ and $\boldsymbol{z}$. Cleaned µPMU data was then fed into the trained DNN during testing to determine $\boldsymbol{x}$, and calculate the maximum testing dataset error.

Step 3: The robustness verification formulation given by (11) was solved. The solution gave the maximum error that the trained DNN will have for an input perturbation that is bounded by $\boldsymbol{\alpha}$. If this solution is greater than the testing error found in Step 2, it means that the robustness formulation has found an input perturbation for which the trained DNN performs worse than what the testing accuracy indicates.

Step 4: The trustworthiness verification formulation given by (12) was solved for every node to determine the smallest input perturbation ($\boldsymbol{\delta}_{\min}$) that is needed to create an error greater than $\beta$ in any of the state estimates. If no perturbation is found for a given node or the smallest perturbation found is greater than the $\boldsymbol{\alpha}$ used in Step 1, then the DNN is deemed trustworthy in the sense that it will not give an error greater than $\beta$ for any input perturbation that is bounded by $\boldsymbol{\alpha}$.

## IV. Results and Discussion

The DNNs created based on the logic proposed in [1] were able to estimate the voltage magnitude and angle of every phase of all the nodes of µPMU-unobservable distribution systems. We used these DNNs to demonstrate the validity of the formulations proposed in Section III. The simulations were performed on a modified IEEE 34-node distribution system (henceforth, called System S1) and a real-world distribution system located in a metropolitan city of U.S. (henceforth, called System S2 [19]). The optimization problems were solved using the branch and bound method in Pyomo coding environment with Gurobi 10.0.1 as the solver on a computer with 384GB RAM, Intel Xeon Platinum 8368 CPU @2.40GHz.

### A. System S1

#### A.1 Robustness Results

System S1 has three distributed generation units having ratings of 135kW, 60kW, and 60kW placed on nodes 822, 848, and 860, respectively. To train a DNN for DSSE, we created a database comprising input, $\boldsymbol{z}$, and output, $\boldsymbol{x}_{ref}$. The database was then split into training and testing datasets. Note that according to the sensor placement algorithm of [1], three µPMUs placed on nodes 800, 850, and 832 of this system are sufficient for performing time-synchronized DSSE using DNNs. A total vector error (TVE) of 0.05% [20] was employed to simulate erroneous voltage and current phasor measurements for these three µPMUs and determine $\boldsymbol{\alpha}$. For a TVE of 0.05%, $\boldsymbol{\alpha}$ was 0.025° and 0.05% for angle and magnitude, respectively. Next, the measurements were normalized and fed as inputs to the DNN, which had two hidden layers with BN, and 30 neurons/layer. Lastly, to prove robustness of the trained DNN for every node of the system, the optimization problem in (11) was solved $2M$ times for each operating condition.

In the first set of simulations, we compared the output of (11) obtained from the trained DNN using the test dataset with the estimation errors produced by the same DNN for the same (test) dataset. Due to space limitation, we only present the results for Phase A voltage magnitude estimation. However, similar observations were made when analysis of magnitudes of other phases as well as angles were conducted. The maximum absolute error of all the test samples is shown in Fig. 4. The blue line shows the maximum absolute error in the output of the DNN for every node where Phase A is present. The orange line shows the maximum absolute error for the same nodes found using (11). It is observed from the figure that for all the nodes the maximum absolute error calculated using the robustness analysis is *greater* than the maximum absolute error calculated from the DNN output. This signifies the importance of robustness analysis for trained DNNs as evaluating the performance of a trained DNN based only on the testing dataset may give more optimistic results (blue line). In addition, the maximum absolute errors found by (11) ensures that as long as the perturbation in the input is less than $\boldsymbol{\alpha}$ ($= 0.05\%$), the error in the state estimates is guaranteed to be less than the values indicated by the orange line of Fig. 4.

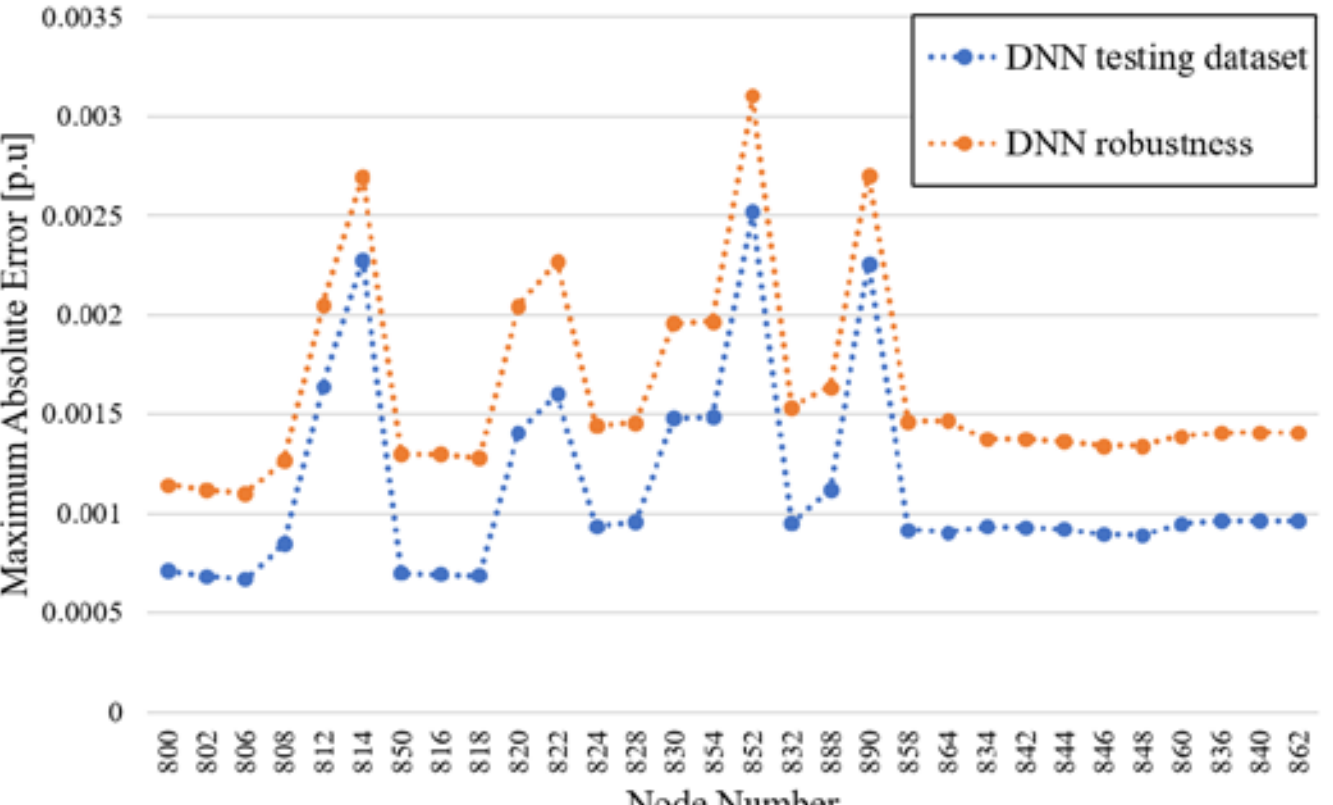

Fig. 4. Comparison of DNN-based voltage magnitude estimation error and DNN robustness analysis for Phase A of System S1.

Next, we tested the sensitivity of the proposed formulation to different sample sizes. The results presented in Fig. 5 are obtained from a dataset of 7,500, 10,000, and 12,500 samples, respectively, each of which is divided into 80% for training and 20% for testing. It can be seen from the figure that the maximum absolute errors found by robustness analysis progressively decrease as the number of samples increase, with the magnitude of the decrease becoming smaller with increase in sample sizes. Since the computational complexity of the proposed approach is a function of the number of samples (and we did not see much improvement after 12,500 samples), we deduced from this analysis that a sample size of 12,500 is sufficient for drawing valid conclusions for this system.

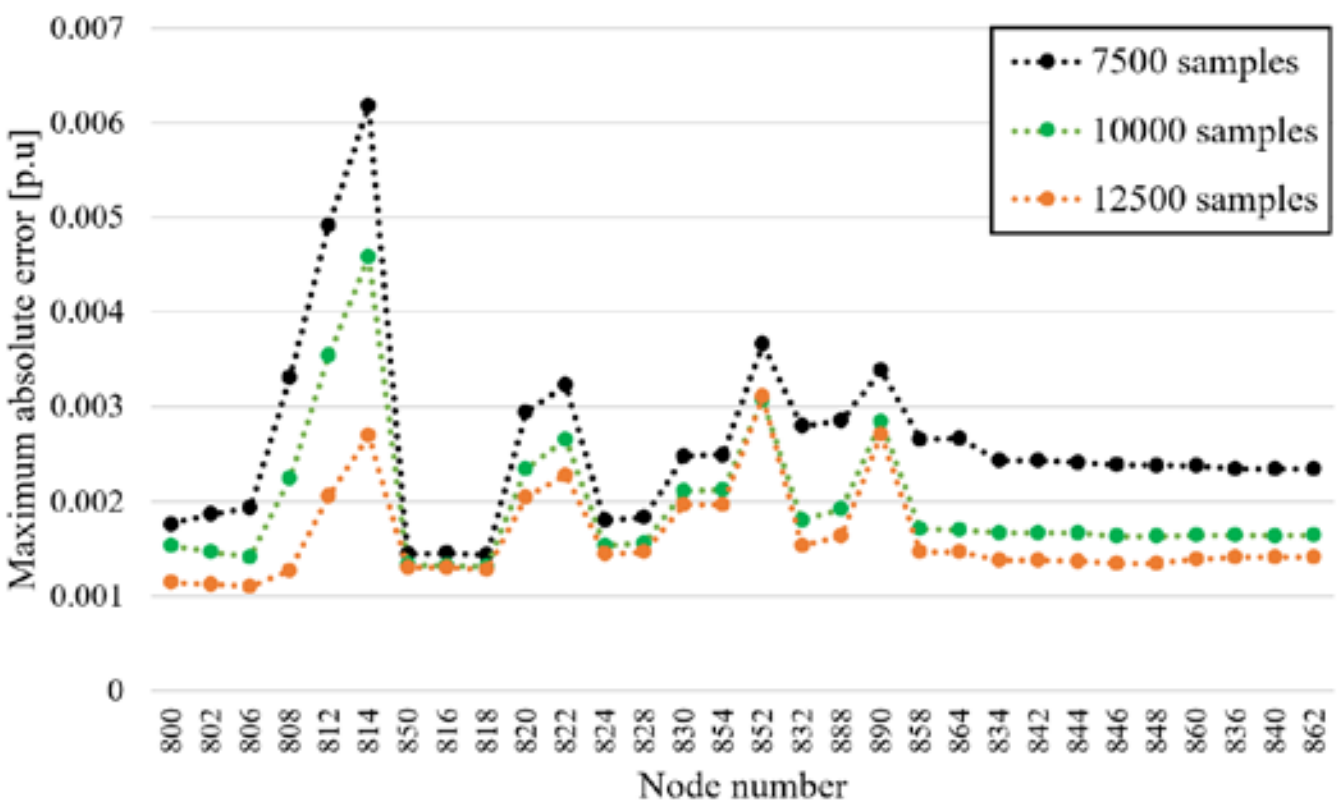

Fig. 5. Robustness analysis for System S1 with different data sizes.

### A.2 Trustworthiness Results

To identify minimum perturbation in μPMU measurements capable of inducing a prespecified error in the state estimates, (12) was employed. We assumed a maximum allowable error of 1% in voltage magnitude estimation, i.e., $\beta = 0.01$. Then, we determined $\boldsymbol{\delta}_{\min}$ specific to each operating condition that resulted in $\beta = 0.01$. To ensure trust in our trained DNN for DSSE, we must verify whether $\boldsymbol{\delta}_{\min}$ consistently exceeds $\boldsymbol{\alpha}$ (= 0.05%). The results obtained are shown in Table I.

TABLE I
TRUSTWORTHINESS RESULTS FOR PHASE A VOLTAGE MAGNITUDE ESTIMATION FOR EACH NODE OF SYSTEM S1.

| Node number | $\boldsymbol{\delta}_{\min}$ [%] | Node number | $\boldsymbol{\delta}_{\min}$ [%] | Node number | $\boldsymbol{\delta}_{\min}$ [%] |
|---|---|---|---|---|---|
| 800 | | 822 | 4.6729 | 864 | 0.9582 |
| 802 | | 824 | | 834 | 0.9667 |
| 806 | | 828 | | 842 | 0.9668 |
| 808 | 10.5745 | 830 | | 844 | 0.9671 |
| 812 | 6.2330 | 854 | | 846 | 0.9665 |
| 814 | 4.7759 | 852 | 6.8931 | 848 | 0.9664 |
| 850 | | 832 | 0.9553 | 860 | 0.9684 |
| 816 | | 888 | 0.9445 | 836 | 0.9691 |
| 818 | | 890 | 0.8169 | 840 | 0.9692 |
| 820 | 3.5278 | 858 | 0.9591 | 862 | 0.9692 |

Table I shows the least amount of input perturbation, expressed as a percentage of the magnitude measurement, that is required to achieve a Phase A voltage magnitude estimation error greater than 1% for every node of System S1. For example, node 814 necessitates a perturbation of at least 4.7759% in the μPMU measurements to generate a 1% error in its Phase A voltage magnitude estimates. For the nodes shown with gray boxes, a $\boldsymbol{\delta}_{\min}$ value that can create a 1% voltage magnitude estimation error in Phase A could not be found, implying that it is not possible to induce an error of $\beta$ (= 0.01) in them. However, it should be noted that by lowering the value of $\beta$, a corresponding $\boldsymbol{\delta}_{\min}$ could be found for these nodes.

The findings presented in Table I instill trust in the trained DNN since for the nodes for which $\boldsymbol{\delta}_{\min}$ was found, it always exceeded the value of $\boldsymbol{\alpha}$ (= 0.05%). It can be further implied from the table that based on the assumed TVE accuracy of the μPMUs, we can be confident that the trained DNN will consistently provide estimates with a voltage magnitude error of no more than 1%. Finally, the results obtained in Fig. 4 and Table I prove the robustness and trustworthiness of the created DNN for System S1 and its ability to give accurate voltage magnitude estimations within the prespecified measurement error bounds.

## B. System S2

### B.1 Robustness Results

In this test system, μPMU measurements were only possible at the feeder-head (see Fig. 6 for a depiction of this system). Note that having real-time measurements only at the feeder-head is common for most distribution systems. Therefore, it was of interest to evaluate performance of time-synchronized DNN-based DSSE in situations where additional μPMU placement cannot be done due to budget constraints. There are 642, 665, and 637 nodes in Phase A, Phase B, and Phase C of this system, respectively, whose voltages must be estimated for different operating conditions. Additionally, this feeder has 766 household/commercial roof-top solar photovoltaic units, implying that it has a high penetration of renewable energy resources. Thus, this was an ideal test system for investigating scalability as well as renewable-rich system handling capability.

Due to the sheer number of nodes in this system we show the difference between robustness analysis and DNN testing dataset for Phase A voltage magnitude estimation as a histogram (see Fig. 7). We display the maximum absolute error found by robustness analysis for the $i^{th}$ node, denoted by $R_i$, and the maximum absolute error based on the testing dataset for the same node, denoted by $T_i$. The histogram shows the numerical difference between $R_i$ and $T_i$, i.e., $R_i - T_i$. The X-axis indicates the ranges of the differences, while the Y-axis denotes the number of nodes belonging to a given range. For example, there are 188 nodes for which the differences in $R_i$ and $T_i$ lie between 0.6×10$^{-4}$ p.u. and 2.6×10$^{-4}$ p.u. It is evident from the histogram that, for all nodes, the difference is *always positive*, as there are zero nodes for which $R_i - T_i$ is less than 0.6×10$^{-4}$ p.u. This indicates that robustness analysis consistently finds adversarial examples that could result in higher errors compared to the testing dataset. Similar observations were made when the analyses were conducted for the other phase magnitudes and angles. This implies that when reporting the accuracy of the trained DNN for DSSE, it is more appropriate to present the robustness analysis results than just the testing dataset results. In summary, the proposed robustness analysis offers a means to provide guarantees for ReLU-based regression DNNs by accounting for the existence of potential errors beyond what is evident from the testing dataset alone.

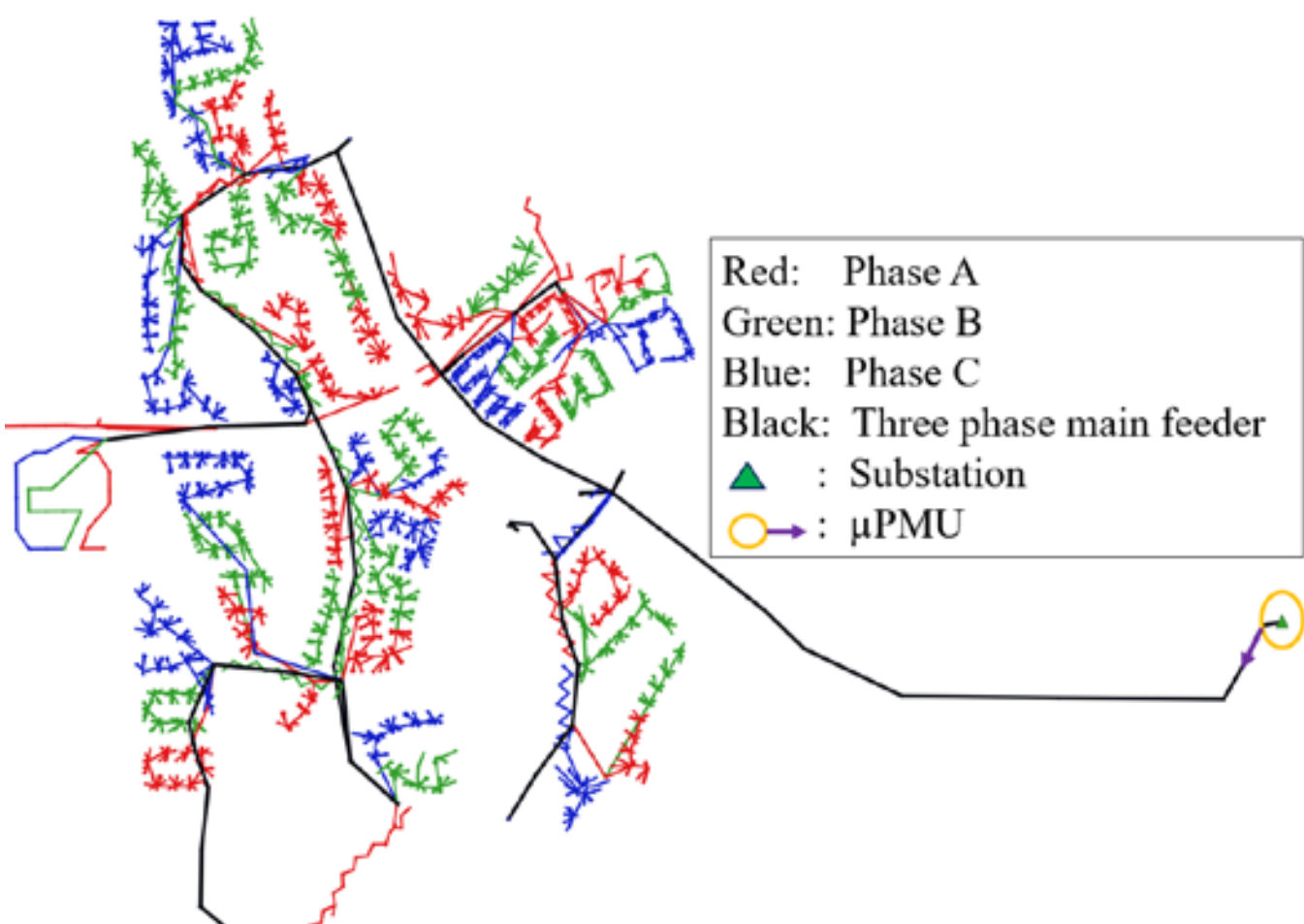

Fig. 6. System S2 with one μPMU available at feeder head.

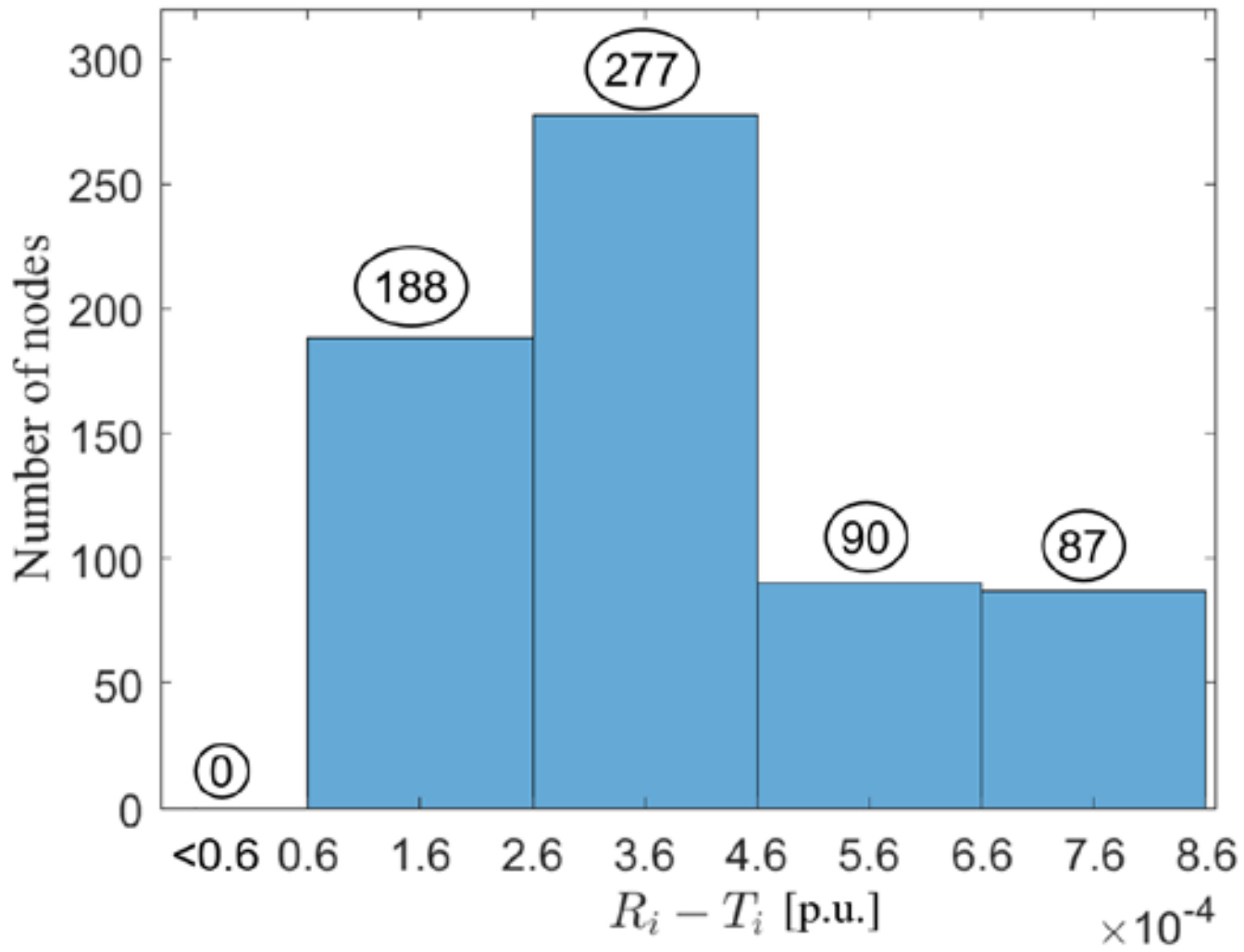

Fig. 7. Difference in maximum absolute errors (in p.u.) obtained using DNN robustness analysis ($R_i$) and DNN-based voltage magnitude estimation error ($T_i$), for all 642 nodes in which Phase A was present in System S2.

### *B.2 Trustworthiness Results*

The results for the trustworthiness analysis for System S2 are presented in the form of a histogram in Fig. 8. Since System S2 is a real-world distribution system that requires reliable operation, we have chosen a smaller value for $\beta$ (= 0.002) to ensure that the estimation error for all nodes and operating conditions never exceed 0.2%. For 527 nodes of this system, no $\boldsymbol{\delta}_{\min}$ was found using (12), similar to the nodes with gray boxes in Table I. As such, they are not included in the figure. Hence, Fig. 8 shows the number of nodes for which a $\boldsymbol{\delta}_{\min}$ was found, as well as the corresponding value intervals. For example, the first non-zero bar indicates that there are 16 nodes in System S2 for which $\boldsymbol{\delta}_{\min}$ lies between 5.4% and 8.6%. The fact that the minimum value in Fig. 8 is 5.4% proves that in order to have a 0.2% error in voltage magnitude estimation, an error of at least 5.4% must be injected into the input data. Since this value is much higher than the $\boldsymbol{\alpha}$ value of 0.05%, this analysis instills trust in our trained DNN for DSSE as it ensures that the estimation error will always be less than 0.2%.

## *C. Discussion*

### *C.1 Strategies to Address Computational Burden*

The proposed approach is built on a MILP-based formulation whose worst-case run-time complexity is exponential. For example, the computational burden of the verification formulations developed in Section III is of the order of $O(S2^{KN})$, where $S$ denotes the total number of samples ($K$ and $N$ have already been defined in Section III.A). Since optimization formulations with exponential time complexity face scalability issues when applied to problems involving large numbers of variables, we have employed three strategies to lower the severity of this issue for the proposed application, namely, time-synchronized DSSE in μPMU-unobservable distribution systems.

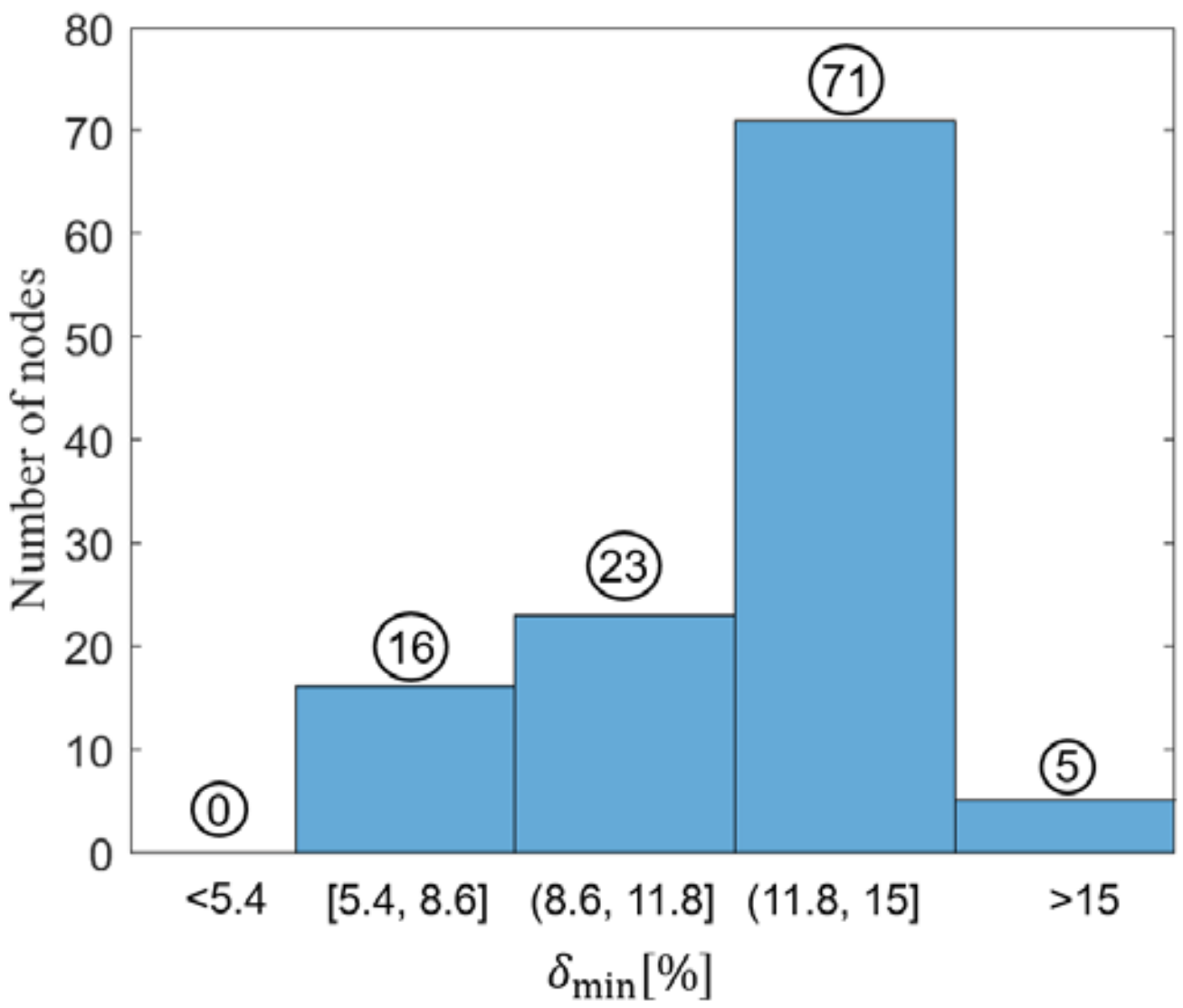

Fig. 8. Trustworthiness results (in %) for Phase A voltage magnitude estimation for 115 nodes of System S2. For the remaining 527 (= 642-115) nodes, no $\boldsymbol{\delta}_{min}$ value was found.

Strategy 1 – Incorporation of BN layer: It was observed that by incorporating BN layers within the DNN, a smaller-sized

DNN could give similar validation accuracy as a larger-sized DNN that did not have BN layers. For example, in the absence of BN, a DNN with eight hidden layers and 500 neurons/layer was needed for System S1 for achieving similar accuracy as the DNN with BN described in Section IV.A. This considerable reduction in the size of the DNN also reduced the number of integer variables in the proposed MILP formulation by a significant amount, resulting in faster convergence of the optimization process.

Strategy 2 – Identifying *always-dead* and *always-active* neurons: During the training process, the output of each neuron was monitored. It was observed that some neurons are *always-active*, while others become *always-dead*, outputting zero. For these neurons, the corresponding binary integer variable, $r_k^n$, was fixed to 1 and 0, respectively, reducing the number of integer variables required for post-training robustness and trustworthiness analyses. This strategy improves the efficacy of the proposed methodology even for large DNNs.

Strategy 3 – Effective parallelization: This was done in two ways. First, the verification formulations were implemented parallelly for the three phases (since DSSE is performed on a per-phase basis). Second, the verification formulations are specific to a given power system node and can be performed independently of any other node. Therefore, different nodes of the test system were grouped together into clusters (e.g., 100 nodes for System S2), and these clusters were solved in parallel. Since both of these ways are agnostic of the DNN size, they can be easily applied to large DNNs for DSSE.

### *C.2 Practical Significance of the Proposed Formulations*

Trained ML models are prone to poor performance due to the presence of adversarial examples that can be present in the input space domain but not seen during training and testing stages. In the context of the DNN-based DSSE application, this can be due to the presence of non-Gaussian noise in the μPMU measurements. Most studies have modeled the noise as a zero-mean Gaussian distribution, but in reality, the noise model could be non-Gaussian [21]. In such a scenario, if a DNN is trained for a Gaussian measurement noise, but is tested on a non-Gaussian measurement noise, then relying on the testing accuracy alone may not give the correct picture (in the worst case, it may give a sense of false security). Moreover, as long as the noise amount is low, it will not be detected/corrected by any bad data detection/correction module.

This is precisely the scenario where the proposed verification formulations become crucial. Consider the robustness verification formulation that calculates the maximum error in the state estimates caused by a perturbation bounded by $\boldsymbol{\alpha}$ in the input measurements. Once this maximum error is found, one can say with certainty that as long as the input is corrupted by a perturbation bounded by $\boldsymbol{\alpha}$ (irrespective of the distribution that the perturbation may have), the DNN-based DSSE error will be less than or equal to this calculated maximum error value. This is a powerful result that clearly indicates the practical significance of the proposed approach for mission-critical systems such as the electric power grid.

## V. CONCLUSION

The black box nature of a DNN often makes power system operators question the validity of the obtained results. This is because although a well-trained DNN can make accurate predictions, it might lack requisite robustness to (adversarial) input perturbations. Therefore, providing formal guarantees of DNN performance is necessary for ensuring their acceptability in the power system. To this end, we formulated two verification problems, namely, robustness and trustworthiness, for DNN-based time-synchronized DSSE using MILP. The robustness formulation finds the maximum error in the output for a given bounded perturbation in the input, while the trustworthiness formulation finds the minimum perturbation in the input that is required to produce a given error in the output. The proposed formulations are also applicable to DNN-based regression problems in other domains.

The analytical verification of DNN-based time-synchronized DSSE was first performed on a modified IEEE 34-node system. It was confirmed that the robustness analysis conducted using the testing data on a DNN resulted in a higher error than what was observed when the same data was fed as an input into that DNN. This implied that relying on the outputs of the testing data alone (i.e., without a robustness analysis) might result in a sense of false security, which is dangerous for mission-critical systems such as power systems. Through trustworthiness analysis, it was observed that we can verify the adherence of the estimation error to a prespecified threshold that is based on the characteristics of the inputs (e.g., permissible error in μPMU measurements). Lastly, the applicability of the proposed method to a real-world, large-scale, renewable-rich distribution system was demonstrated, confirming its practical utility. A future scope of this work will be to address the exponential run-time complexity of the proposed formulations by creating verification problems that do not involve MILP.

## DISCLAIMER

**Behrouz Azimian** (Student Member, IEEE) received the Bachelor of Science degree in electrical engineering from Iran University of Science and Technology, Tehran, Iran, in 2016. He Received the Master of Science degree in electrical engineering from Alfred University, NY, USA, in 2019. He is now a Ph.D. student at Arizona State University. His research interests include machine learning, deep learning, application of artificial intelligence and optimization methods in power system problems such as state estimation, electricity markets, and electric vehicle charging scheduling.

**Shiva Moshtagh** (Student Member, IEEE) holds a B.Sc. degree in electrical engineering from Jundi-Shapur University of Technology, Dezful, Iran, which she earned in 2015. She received her M.Sc. degree in electrical engineering from Imam Khomeini International University, Qazvin, Iran, in 2019. Currently, she is pursuing a Ph.D. degree in electrical engineering at Arizona State University in Tempe, AZ, USA. Her research interests primarily revolve around the application of machine learning techniques, including deep neural networks and graph neural networks, to address challenges within the field of power systems. Her work is focused on enhancing the efficiency and reliability of power systems through artificial intelligence-driven solutions.

**Anamitra Pal** (Senior Member, IEEE) received the B.E. degree (summa cum laude) in electrical and electronics engineering from the Birla Institute of Technology at Mesra, Ranchi, India, in 2008, and the M.S. and Ph.D. degrees in electrical engineering from Virginia Tech, Blacksburg, VA, USA, in 2012 and 2014, respectively. He is currently an Associate Professor in the School of Electrical, Computer, and Energy Engineering at Arizona State University (ASU), Tempe, AZ, USA. His research interests include data analytics with a special emphasis on time-synchronized measurements, artificial intelligence-applications in power systems, renewable generation integration studies, and critical infrastructure resilience. Dr. Pal has received the 2018 Young CRITIS Award for his contributions to the field of critical infrastructure protection, the 2019 Outstanding Young Professional Award from the IEEE Phoenix Section, the National Science Foundation CAREER Award in 2022, and the 2023 Centennial Professorship Award from ASU.

**Shanshan Ma** (Member, IEEE) received the M.S. degree from the Department of Electrical Engineering and Computer Science, South Dakota State University, Brookings, SD, USA, in 2015, and the Ph.D. degree from the Department of Electrical and Computer Engineering, Iowa State University, Ames, IA, USA, in 2020. She served as a Postdoctoral Research Scholar at the School of Electrical, Computer and Energy Engineering at Arizona State University, Tempe, from 2020 to 2023. She is currently working as a Principal Engineer at Quanta Technology, Raleigh, NC, USA. Her current research interests include optimization and control in power distribution systems.